\documentclass[11pt]{article}
\usepackage{acl}
\usepackage{times}
\usepackage{graphicx} 
\usepackage{multirow}
\usepackage{float}
\title{Analyzing the Sensitivity of Vision Language Models in Visual Question Answering}

\author{Monika Shah, Sudarshan Balaji, Somdeb Sarkhel, Sanorita Dey, Deepak Venugopal \\
University of Memphis, TN, USA \\
Adobe Research, San Jose, CA, USA \\
University of Maryland Baltimore County, USA \\ \{mshah2, sbalaji, dvngopal\}@memphis.edu, sarkhel@adobe.com, sanorita@umbc.edu}
  
\date{}
\newcommand{\eat}[1]{}

\begin{document}

\maketitle
\begin{abstract}
    We can think of Visual Question Answering as a (multimodal) conversation between a human and an AI system. Here, we explore the sensitivity of Vision Language Models (VLMs) through the lens of cooperative principles of conversation proposed by Grice. Specifically, even when Grice's maxims of conversation are flouted, humans typically do not have much difficulty in understanding the conversation even though it requires more cognitive effort. Here, we study if VLMs are capable of handling violations to Grice's maxims in a manner that is similar to humans. Specifically, we add modifiers to human-crafted questions and analyze the response of VLMs to these modifiers. We use three state-of-the-art VLMs in our study, namely, GPT-4o, Claude-3.5-Sonnet and Gemini-1.5-Flash on questions from the VQA v2.0 dataset. Our initial results seem to indicate that the performance of VLMs consistently diminish with the addition of modifiers which indicates our approach as a promising direction to understand the limitations of VLMs.
\end{abstract}
\section{Introduction}

Vision Language Models (VLMs)~\cite{team2024gemini,liu2023visual,hurst2024gpt} that unify Large Language Models with computer vision have made significant advances in multimodal tasks such as image captioning~\cite{yang2019auto, cornia2020meshed, wang2022end} and visual question answering (VQA)~\cite{antol2015vqa}. However, we are just beginning to understand the reasoning capabilities and more importantly, the limitations of these models~\cite{campbell2024understanding}. In this work, inspired by theories from cognitive science, we understand the behavior of VLMs in VQA when we {\it increase the cognitive load in comprehending questions}. Specifically, in Grice's classical theory of {\it cooperative principles}~\cite{grice1975logic}, it is known that humans acting cooperatively in a conversation typically need to follow a set of rules commonly known as {\it Grice's maxims}. These maxims make conversation more effective and ensure efficient communication. However, it is known from previous studies that even when these maxims are violated, humans can comprehend conversation easily~\cite{davies2000grice}.  However, violations to Grice's maxims places greater cognitive burden on the listener~\cite{jacquet2018gricean}. 


In this work, we study how VLMs react when Grice's maxims are violated. Specifically, we treat VQA as a single utterance conversation where a human is asking the AI model a question to which the AI model responds with an answer. We introduce modifiers into human-crafted questions
that adds greater detail to a question. At the same time, these details typically tend to violate Grice's maxims since they were not deemed to be essential when a human crafted the original question. While an AI model could benefit from the added information, processing modifiers will increase the reasoning required to answer the question. We add two types of modifiers, namely, {\it visual} and {\it relational} modifiers. The visual modifiers add more detail related to visual properties such as color, shape, etc., while relational modifiers add details related to spatial relationships. 

We use VLMs to generate a modified question with either visual or relational modifiers. Next, we verify if the modified question changes human perception. That is, if humans can answer the modified question with an answer that is equivalent to the answer to an unmodified question, this implies that the modifier does not alter the question. Therefore, we would expect a VLM to be able to do a similar type of reasoning. We evaluate this on three state-of-the-art VLMs, {\it GPT-4o}~\cite{OpenAI2024}, {\it Gemini-1.5-Flash}~\cite{team2024gemini} and {\it Claude-3.5-Sonnet}~\cite{Anthropic2024} on the VQA v2.0 dataset. That is, we generate modified questions from each of these VLMs and evaluate the responses of each VLM to the modified questions. Our initial results seem to indicate that VLMs are sensitive to modifications to questions. In particular, we find that there is a consistent performance degradation in the presence of modifiers. In particular, when modifiers are added through Gemini-1.5-Flash, the performance degradation is more significant in all 3 VLMs.

\section{Related Work}


Following the original VQA task~\cite{antol2015vqa}, several improved datasets for VQA have been developed~\cite{goyal2017making,selvaraju2020squinting,tan2019lxmert} to evaluate VQA systems. More recently, the trend has shifted towards incorporating LLMs within the evaluation process. For instance, ~\cite{zhou2023lima} uses ChatGPT to automatically evaluate
model outputs on a Likert scale. The work in ~\cite{manas2024improving} leverages LLMs to evaluate answers. Specifically, it formulates VQA as an answer-rating task where the LLM (Flan-T5~\cite{chung2024scaling}, Vicuna-v1.3~\cite{chiang2023vicuna} and GPT-3.5-Turbo) is instructed to score
the correctness of a candidate answer given a set of reference answers. The work in ~\cite{britton2022question} is related to our approach where it adds question modifiers to VQA and analyzes its effect on LXMERT~\cite{tan2019lxmert}. However, there has not been a significant amount of work that relates the reasoning of VLMs in VQA grounded in principles of human cognition which is the direction we follow in this work.

\section{Pragmatics in Visual Question Answering}



\begin{figure}[t]
\includegraphics[width=\columnwidth, trim=90 120 100 50, clip]{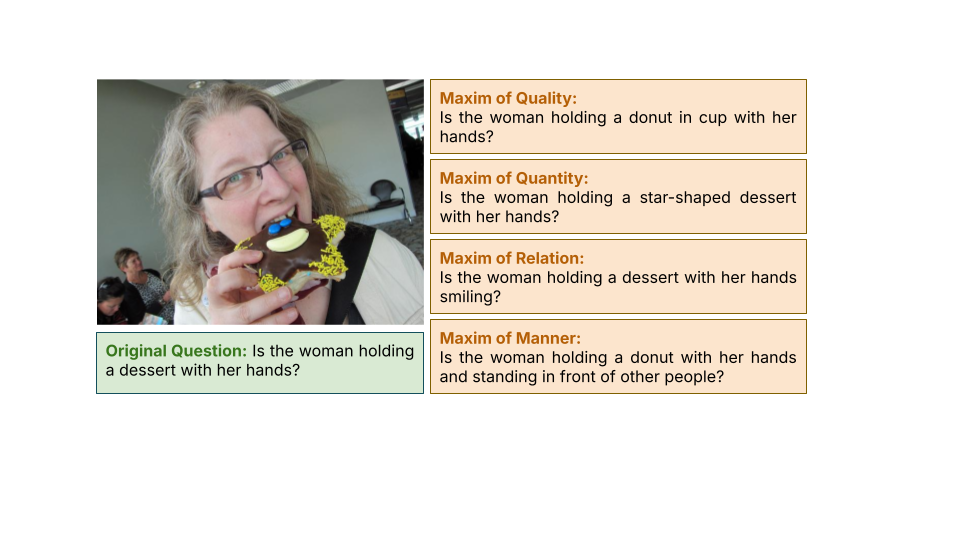}
    \caption{Original question of the green is the question that satisfies the Grice's maxim and the questions with modifiers that violates the Grice's maxim.}
    \label{fig:griceexample}
\end{figure}

Grice's classical theory of cooperative principles in pragmatics is widely used to characterize human conversation. Specifically, Grice developed principles that explain effective conversation between participants assuming that the participants have a common goal of understanding each other and therefore act cooperatively. These principles are summarized in four maxims, namely, the maxim of quality, quantity, relation and manner. The maximum of quality suggests that speakers should be as truthful as possible and only say what they believe to be true based on evidence. The maxim of quantity suggests that the right amount of information must be provided in a conversation, i.e., one should not add too much or too little information. The maxim of relation suggests that a speaker should stay relevant to the topic and the maxim of manner suggests the need to avoid ambiguity and focus on clarity.

While Grice's maxims characterize effective conversation, violation of Grice's maxims does not mean that the conversation is incomprehensible ~\cite{davies2000grice}. Specifically, since participants are assumed to be acting cooperatively, if a speaker violates a maxim, then the burden of understanding falls on the listener. That is, the listener is expected to work harder (cognitively) to comprehend the intention behind utterances that violate the maxims. We use this principle as a way to understand the limitations of VLMs. Specifically, we think of the VQA task as a conversation that involves a single utterance between two participants, i.e., one participant is a human who asks a question to the AI model and the other participant is the AI model that needs to generate an answer. In cases where the human participant flouts Grice's maxims, there is an increased burden of understanding on the AI model to produce an answer that the human can agree on.

\begin{figure*}[htbp]
    \includegraphics[scale=0.45]{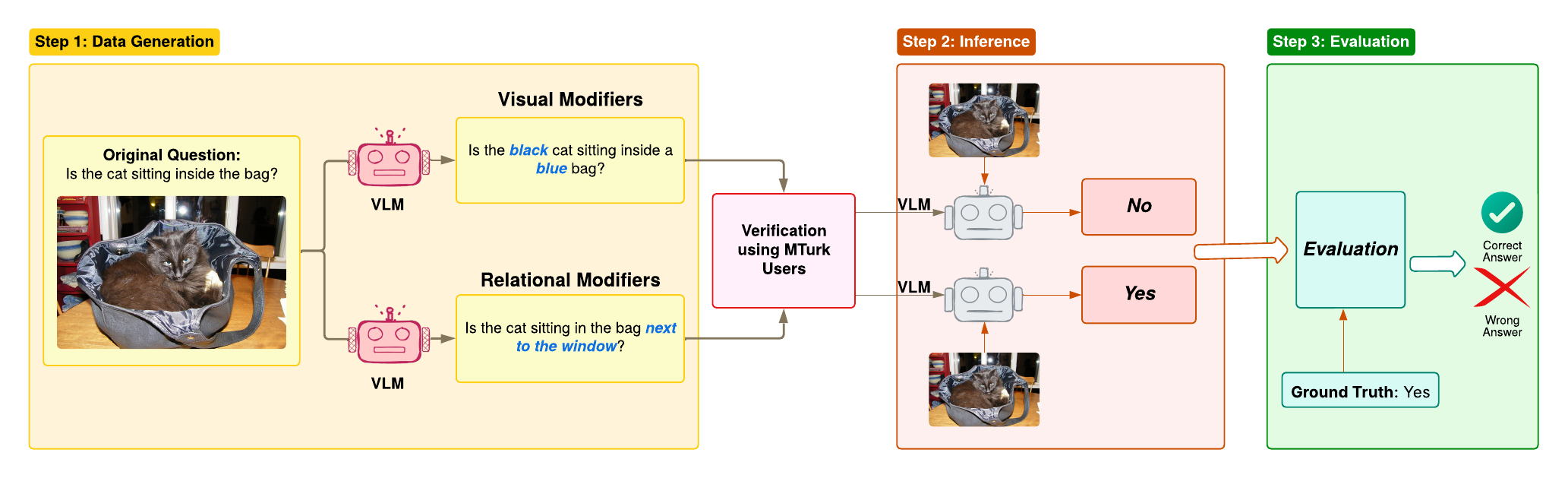}
    \caption{Illustrating our workflow. We generate modified questions from human-crafted questions using a VLM. Next, we verify if the modifier changes human perception of the question by comparing answers to the modified questions (collected through AMT) to the answers given to the original questions. For questions where the answers are alike, we evaluate if a VLM gives similar answers to the original and modified questions.}
    \label{fig:process}
\end{figure*}

\subsection{Adding Modifiers to VQA}

Fundamentally, the purpose of modifiers in text is to add more detail. Modifiers can add more specifics to a description, clarify information to improve comprehensibility or can make text more engaging for a reader. At the same time, from a cognitive perspective, processing modifiers places greater demands on attention and reasoning capabilities. Specifically, it is known that to understand text with greater syntactic complexity (which can occur if modifiers are added to questions) the level of neural activity in the brain increases~\cite{just1996brain}. Further, if we consider our view of VQA as a conversation between a human and an AI model, adding modifiers to a human-crafted question is very likely to violate Grice's maxims which again results in the need for greater reasoning capability.
For instance, consider the example shown in Fig.~\ref{fig:griceexample}. The original question written by a human seems to follow Grice's maxims. However, by adding modifiers, we violate these maxims as illustrated in the example. Importantly though, each of the modified questions can be easily answered by humans even when they violate at least one of the maxims and have increased complexity of the question (for instance, in our AMT study, humans answered modified questions with answers similar to those in unmodified questions).
Pragmatically, since the AI is interacting with humans (e.g.in standard VQA, we use human-generated questions~\cite{antol2015vqa}), such an interaction is likely to follow Grice's maxims assuming that humans are acting cooperatively and not maliciously. That is, if we consider the example shown in Fig.~\ref{fig:griceexample}, it is unlikely that a human would ask the AI any of the questions where the modifiers violate the maxims. However, human reasoning is fairly robust to such modifications. Our goal is use these modifiers to help us explain if the reasoning mechanism of the model is equally robust.
Specifically, the modifiers may i) describe new concepts such as the {\em star-shape} that describes the shape of the dessert,  ii) add additional information such as where the woman is standing or iii) add ambiguity such as if the woman's facial expression describes a smile. The AI model could in theory use the additional context to improve the accuracy of its answers in the VQA task. In other cases, the accuracy may diminish either due to increased ambiguity or a lack of model capacity to process modifiers.

\subsection{Evaluation Methodology}

We add modifiers to human-written questions targeting specific properties in the image. Specifically, here, we consider two properties that are broad enough to describe the scene in an image in greater detail, i.e., {\em visual properties} and {\em relational properties}. Specifically, visual properties refer to attributes such as color, shape, texture, etc. for objects that are observed in the scene. Relational properties refer to spatial relationships in the scene such as  {\em next to}, {\em on top of}, etc.
We prompt a VLM (with the image and original question) to generate the modified question with a specific type of modification (i.e., visual/relational). We instruct it to add the modifier without changing the answer to the original question and also without altering the question type (e.g. a {\em what} question needs to remain a {\em what} question). Further, we also instruct it to not alter the context of the question significantly. 
Next, we use Amazon Mechanical Turk (AMT) to verify if violations to Grice's maxims alter human perception.
Specifically, we ask a human to answer a question with a modifier and compare this answer to answers given to the original question. Note that for questions where the original answer has a {\em unique ground truth} (yes/no questions and numeric questions), it is easy to verify if the answer changes from the original answer. However, for a question that is open-ended, there could be multiple ground truth answers. For such cases, we use an LLM to compare answers to the modified and unmodified questions, and instruct it to quantify the similarity between them on a discrete 1-10 scale. An illustration of our evaluation workflow is shown in Fig.~\ref{fig:process}. More details about the prompts and the AMT study are presented in the appendix.

\begin{table*}
  \centering
  \resizebox{1.75\columnwidth}{!}{%
  \begin{tabular}{|l|l|l|l|l|l|l|}
    \hline
    \multirow{2}{*}{\bf Model / Modifier} &
      \multicolumn{2}{c|}{GPT-4o} &
      \multicolumn{2}{c|}{Gemini-1.5-Flash} &
      \multicolumn{2}{c|}{Claude-3.5-Sonnet} \\
      
    & \multicolumn{1}{c}{\it Visual} & \multicolumn{1}{c|}{\it Relational} & \multicolumn{1}{c}{\it Visual} & \multicolumn{1}{c|}{\it Relational} & \multicolumn{1}{c}{\it Visual} & \multicolumn{1}{c|}{\it Relational} \\
    \hline
    GPT-4o & 1.06\% & 2.91\% & \textcolor{red}{\bf 8.22\%} & \textcolor{blue}{\bf 8.22\%} & -0.26\% & 3.18\% \\
    \hline
    Gemini-1.5-Flash & 8.71\% & 7.08\% & \textcolor{red}{\bf 11.44\%} & \textcolor{blue}{\bf 13.07\%} & 5.17\% & 6.26\% \\
    \hline
    Claude-3.5-Sonnet & 4.86\% & 3.78\% & \textcolor{red}{\bf 8.91\%} & \textcolor{red}{\bf 6.21\%} & 1.35\% & 3.51\% \\
    \hline
  \end{tabular}
  }
  \caption{$\%$ change in accuracy for questions with yes/no answers and numeric answers (larger values indicate the model performed worse on modified questions). The column headings indicate which VLM was used to generate modified questions and the row headings indicate the VLM we are evaluating. The values in \textcolor{red}{red} show the worst performing VLM model/modifier combination when adding visual modifiers and the values in \textcolor{blue}{blue} show the worst performing model/modifier combination for relational modifiers.}
  \label{tab:yesno}
\end{table*}

\begin{table*}
  \centering
  \resizebox{1.75\columnwidth}{!}{%
  \begin{tabular}{|l|l|l|l|l|l|l|}
    \hline
    \multirow{2}{*}{\bf Model / Modifier} &
      \multicolumn{2}{c|}{GPT-4o} &
      \multicolumn{2}{c|}{Gemini-1.5-Flash} &
      \multicolumn{2}{c|}{Claude-3.5-Sonnet} \\
        & \multicolumn{1}{c}{\it Visual} & \multicolumn{1}{c|}{\it Relational} & \multicolumn{1}{c}{\it Visual} & \multicolumn{1}{c|}{\it Relational} & \multicolumn{1}{c}{\it Visual} & \multicolumn{1}{c|}{\it Relational} \\
    \hline
    GPT-4o & 4.58\% & \textcolor{blue}{\bf 6.87\%} & \textcolor{red}{\bf 8.10\%} & 6.08\% & 1.96\% & 4.54\% \\
    \hline
    Gemini-1.5-Flash & 5.82\% & 4.91\% & \textcolor{red}{\bf 8.13\%} & \textcolor{blue}{\bf 6.59\%} & 3.43\% & 6.31\% \\
    \hline
    Claude-3.5-Sonnet & 6.12\% & 5.96\% & \textcolor{red}{\bf 8.72\%} & \textcolor{blue}{\bf 8.32\%} & 3.89\% & 7.16\% \\
    \hline
  \end{tabular}
  }
  \caption{$\%$ change in accuracy for questions with open-ended answers (larger values indicate the model performed worse on modified questions). The column headings indicate which VLM was used to generate modified questions and the row headings indicate the VLM we are evaluating. The values in \textcolor{red}{red} show the worst performing VLM model/modifier combination when adding visual modifiers and the values in \textcolor{blue}{blue} show the worst performing model/modifier combination for relational modifiers.}
  \label{tab:openended}
\end{table*}



\subsubsection{Results}

We evaluate 3 well-known VLMs, {\it GPT-4o}, {\it Gemini-1.5-Flash} and {\it Claude-3.5-Sonnet} using the VQA v2.0 dataset ~\cite{goyal2017making}.  We added visual and relational modifiers to 1000 questions from the test set of VQA v2.0. We selected these questions such that we had an equal number of instances corresponding to each question type (there are 55 question types, e.g. {\it what}, {\it why}, {\it is}, {\it how}, etc.). The sampled data we used consists of 500 yes/no and numeric questions (where the answer is a number) and 500 open-ended questions. We evaluated each VLM on modified questions generated from each of the 3 VLMs. 


Tables~\ref{tab:yesno}, ~\ref{tab:openended} show the $\%$ change in accuracies of answers given by the VLM when modifiers are added to the original questions. The results in Table~\ref{tab:yesno} correspond to yes/no and numeric questions where we can evaluate the answers exactly since these questions have a unique ground truth answer.
As seen from our results, the positive values of $\%$ change in almost all cases indicates that the models performed worse on modified questions regardless of which VLM performed the modification. Modifiers added by Gemini-1.5-Flash seemed to be the hardest ones to process for all 3 VLMs since the average $\%$ change over all the VLMs was the largest. The modifiers added by Claude-3.5-Sonnet seemed to be easier to process for all 3 VLMs since the average $\%$ change in accuracy was the smallest across all the models. This seems to indicate that Claude-3.5-Sonnet may not add substantially detailed modifiers compared to the other models. Interestingly, Gemini-1.5-Flash performed worse with self-modified questions compared to modifications by other VLMs both for visual and relational modifiers. In the case of GPT-4o, self-modified questions did not result in a significant change to the model's accuracy as compared to its change in accuracy on questions modified by other VLMs.  This indicates that GPT-4o can handle specific forms of modifications which is built into its prior but struggles with other forms of modifications. While our results indicate that some VLMs perform better than others, the specific reasons for why this may be the case is still unclear. We plan to explore this in future.


For the open-ended question results shown in Table~\ref{tab:openended}, we use GPT-4o to evaluate the similarity between human-generated answers to the original question (which we collected using AMT) and the answers given by the model to the original and modified questions. In this case, we only compare the texts using GPT-4o. For each question, we used answers from 3 AMT workers and considered all the 3 similarity scores provided by GPT-4o on a discrete scale between 1 and 10. Table~\ref{tab:openended} shows the $\%$ difference between these scores for answers given by the VLM to the original questions with those given by the VLM for modified questions. Overall, similar to our earlier result, in all cases, the models performed worse on modified questions (positive $\%$ change values) regardless of which VLM performed the modification. Further, consistent with our results on yes/no and numeric questions, modifications by Gemini-1.5-Flash were the hardest to process (largest average $\%$ change) for all 3 VLMs while Claude-3.5-Sonnet modifications were the easiest to process (smallest average $\%$ change). There was no consistent pattern to indicate whether the models performed better/worse on modifications of open-ended questions compared to the yes/no, numeric questions. However since the open-ended questions are scored approximately, the results in Tables~\ref{tab:yesno} and ~\ref{tab:openended} may not be directly comparable.

\paragraph{Significance tests.} We use a paired test to evaluate if the response of a VLM changes significantly when a modifier is added. Specifically, for yes/no and numeric questions since the answer can be compared exactly with the ground truth to obtain a binary outcome, we use the {\it McNemar's test}. The McNemar's exact test is used to evaluate if there is a significant difference in a dichotomous dependent variable between two  groups. It is used frequently to evaluate drug effects~\cite{trajman2008mcnemar}, and has been shown to have low type I error~\cite{dietterich1998approximate}. To run this test, we pair binary outcomes obtained by comparing the VLM's answer prior to and after question modification with the ground truth. 


Our results showed that in most cases there was significant change in the VLM response ($p<0.05$). However, when the modifiers were added using Claude-3.5-Sonnet, the change in responses of Claude-3.5-Sonnet/GPT-4o was insignificant ($p\geq 0.05$) which again indicates that Claude-3.5-Sonnet may be limited in its ability to add detailed modifiers. The responses of GPT-4o did not significantly change on self-modified questions ($p\geq 0.05$) with yes/no or numeric answers which again may indicate that GTP-4o performs well only when it has a strong prior on the type of modification. One alternate possible explanation is that perhaps GPT-4o stores the context in our interaction with it when generating modified questions and this somehow could influence its response to the modified questions (though we used a separate session for generating modified questions).

For open-ended questions, since the comparison between the ground truth and a VLM's answer does not yield a dichotomous value, we use the {\it Wilcox signed-rank} test (since the data was not normally distributed) instead of the McNemar's test. The results were very similar our findings with the McNemar's test. Claude-3.5-Sonnet/GPT-4o showed no significant change in responses ($p\geq 0.05$) when the modifier was Claude-3.5-Sonnet, and GPT-4o had insignificant change when answering self-modified questions. We plan to further investigate if there are specific linguistic characteristics of the modifiers that makes a question either harder or easier to answer.
\section{Conclusion}

In this work, we studied if VLMs are sensitive to modifications to questions in VQA. Specifically, adding modifiers increases details in a question, but when viewed from the perspective of cooperative principles, they can violate Grice's maxims. Humans can accurately ignore irrelevant details to answer questions even with these violations. We studied if VLMs could do the same in VQA by generating modified questions from human-crafted questions that preserve the original answer. We used 3 state-of-the-art VLMs in our study and showed that in most cases, adding modifiers to questions degrades the performance of the VLM. Based on these initial results, we plan to develop more detailed experiments to understand the types of modifications that VLMs are better at processing. Further, while our current results reveal that the performance of VLMs drops in the presence of modifiers, it is not yet clear as to why such a drop occurs. In future work, we plan to analyze the reasons for why some VLMs tend to perform more poorly than others in modified questions.

\section{Limitations}

Following are the limitations associated with this work.
\begin{enumerate}
    \item This work assumes that human-written questions follow Grice's maxims of conversation. However, it may be the case that since humans are asking an AI a question (as opposed to talking to fellow humans), some of these maxims are violated even in human-generated questions.
    \item Since the internal details of how VLMs handle prompts are not clearly known, there could be some bias associated with self-modified questions. That is, if a VLM tries to answer its own modified question since it would have access to the previous prompts (instructing it to add modifiers), it may be able use it in the response to modified questions. Even though, we provided the modification as a separate prompt, there could be some bias in the results of self-modified questions if the prompts are not completely independent.
    \item Since open-ended questions do not have a unique ground truth answer, the evaluation we used may have a bias compared to those which have a unique ground truth answer.
\end{enumerate}

\section*{Acknowledgments}
This research was supported by NSF award $\#$2008812, and awards from the Gates Foundation and Adobe. The opinions, findings, and results are solely the authors' and do not reflect those of the funding agencies.

\bibliography{custom}
\section*{Appendix A: VLM Prompts}

\textit{Prompt to generate modified questions targeting visual properties}:

\noindent Instruction:
Your task is to generate 1 different modified version of the original question about an image, ensuring that each modification preserves the original answer from the provided question and provide the type of the visual attribute that was added to the original question.
 
Given an image and its original question, create 1 unique modification by adding different types of visual attributes to the objects in the original question. The visual attributes can be of the following types:
\begin{itemize}
    \item Physical properties (size, color, shape etc.) of the object
    \item Appearance characteristics (texture, pattern etc.) of the object
    \item Visual state (new, old, clean, dirty etc.) of the object
\end{itemize}
NOTE: You are not limited to the categories mentioned above. You are free to categorize as you see fit.
 
\noindent IMPORTANT: When adding visual attributes to questions, ensure that your modifications don't inadvertently reveal or hint at the correct answer.
 
\noindent **For the visual attribute categories, please use clear, specific labels such as:
\begin{itemize}
    \item Color (when referring to color attributes)
    \item Texture (when referring to surface qualities)
    \item Size (when referring to dimensions)
    \item Shape (when referring to form)
    \item Pattern (when referring to visual arrangements)
    \item Visual state (when referring to condition)
    \item Physical property (when referring to other physical characteristics)
\end{itemize}
This helps maintain consistency in your categorization.**
 
Rules:
Each modification MUST:
    \begin{itemize}
    \item  Preserve the core meaning of the original question
    \item  Yield the same answer as the original question
    \item  Be distinctly different from other modifications
    \item  Use natural, grammatically correct language
    \end{itemize}
 
Avoid:
    \begin{itemize}
    \item  Repeating the same modifier type across the 3 versions
    \item  Making assumptions about details not visible in the image
    \item  Changing the fundamental subject or action in the question
    \end{itemize}

Output:
Modified Questions [LIST]: [Question1]
**Visual attribute [LIST]: [category1]**
 
Example 1:
Original Question: Is the dog skateboarding?
Modified Question [LIST]: [Is the small dog skateboarding?]
Visual attribute [LIST]: [size]
 
Example 2:
Original Question: Is there graffiti shown on the concrete wall?
Modified Question [LIST]: [Is there colorful graffiti shown on the concrete wall?]
Visual attribute [LIST]: [color]
 
IMPORTANT: When adding visual attributes to questions, ensure that your modifications don't inadvertently reveal or hint at the correct answer. The visual attributes should add detail without changing the difficulty level of the question or providing clues that make the answer obvious.\\

\noindent \textit{Prompt to generate modified questions targeting relational properties}:

\noindent Your Task: Generate 1 different modified version of the provided question, ensuring that each modification uses a different relational modifier (positional relationships, for e.g. in front of, on, next to, in, etc.) while preserving the original answer.
 
Instruction:
Given an original question, create 1 unique modification by adding different relational modifiers to the objects in the original question. Each modification must preserve the core meaning and yield the same answer as the original question.
 
Rules:
 
Each modification MUST:
\begin{itemize}
    \item  Use a different relational modifier (e.g., on, under, in front of, next to, in, among, etc.)
    \item  Preserve the core meaning of the original question
    \item  Yield the same answer as the original question
    \item  Be distinctly different from other modifications
    \item  Use natural, grammatically correct language
\end{itemize}

Avoid:
\begin{itemize}
    \item  Changing the fundamental subject or action in the question
    \item  Making assumptions about details not provided in the original question
    \item  Using non-relational modifiers (like color, size, shape, etc.)
 \end{itemize}

Output:
 
Modified Questions [LIST]: [Question1]
Relational Modifier [LIST]: [Modifier1]

Example:
Original Question: Is the dog skateboarding?
Modified Question [LIST]: [Is the dog skateboarding on the sidewalk?]
Relational Modifier [LIST]: [on the sidewalk]

NOTE: DO NOT CHANGE THE MAIN CONTENT IN THE QUESTION. When adding relational attributes to questions, ensure that your modifications don't inadvertently reveal or hint at the correct answer. The relational attributes should add detail without changing the difficulty level of the question or providing clues that make the answer obvious.

\eat{
\noindent\textit{Prompt for open ended answer evaluation}:

You are given two words. Can you tell me how similar they are on a scale of 1 to 10? Provide the integer score only no description.
}

\section*{Appendix B: AMT Details for verification}
We used three workers to answer each question. 
Following is the instruction provided to AMT users to verify the modified questions generated by LLMs;

Instruction: You will see an image and two questions; Q1 (Original Question) and Q2 (Modified Question). The answer for Q1 is shown. Is the same answer correct for Q2?

\begin{figure}[H]
    \centering \includegraphics[width=\columnwidth]{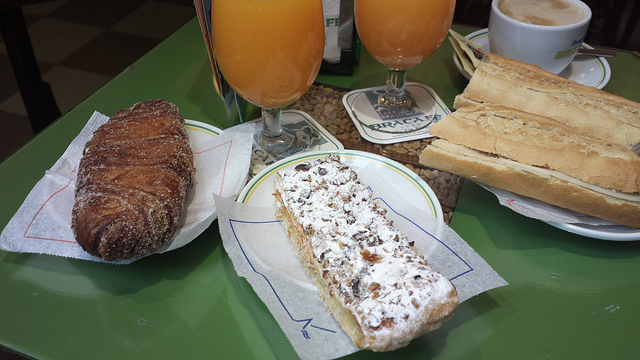}
    \label{fig:exmp1}
\end{figure}

\begin{enumerate}
\item [Q1:] Is there a coffee cup?\\
    Answer: Yes \\ 
\item [Q2:] Is there a white coffee cup? \\
   Answer: Yes \\

Select one of these options:\\
    $ \bigcirc $ Correct Answer \\
    $ \bigcirc $ Incorrect Answer \\
    $ \bigcirc $ Answer is incorrect  in both Q1 and Q2
\end{enumerate}

We consider the modified questions that has the same answer or {\em correct} response from AMT users as the verified questions.

\end{document}